\documentclass{ecai}
\usepackage{graphicx}
\usepackage{latexsym}
\usepackage{times}
\usepackage{latexsym}
\usepackage{amssymb}
\usepackage{amsmath}
\usepackage{multirow}

\begin{document}

\begin{frontmatter}

\title{Hierarchical Text Classification using Contrastive Learning Informed Path Guided Hierarchy}

 \author[A]{\fnms{Neeraj}~\snm{Agrawal}\orcid{0000-0003-1496-6618}\thanks{Corresponding Author. Email: neeraj.agrawal@walmart.com}}
 \author[A]{\fnms{Saurabh}~\snm{Kumar}\orcid{0009-0006-5396-2723}}
 \author[A]{\fnms{Priyanka}~\snm{Bhatt}}\orcid{0009-0009-6431-0490}
  \author[A]{\fnms{Tanishka}~\snm{Agarwal}\orcid{0009-0007-8583-1653}}

\address[A]{Walmart Global Tech, Bangalore, India}

\begin{abstract}
Hierarchical Text Classification (HTC) has recently gained traction given the ability to handle complex label hierarchy. This has found applications in domains like E- commerce, customer care and medicine industry among other real-world applications. Existing HTC models either encode label hierarchy separately and mix it with text encoding or guide the label hierarchy structure in the text encoder. Both approaches capture different characteristics of label hierarchy and are complementary to each other. In this paper, we propose a Hierarchical Text Classification using Contrastive Learning Informed Path guided hierarchy (HTC-CLIP), which learns hierarchy-aware text representation and text informed path guided hierarchy representation using contrastive learning. During the training of HTC-CLIP, we learn two different sets of class probabilities distributions and during inference, we use the pooled output of both probabilities for each class to get the best of both representations. Our results show that the two previous approaches can be effectively combined into one architecture to achieve improved performance. Tests on two public benchmark datasets showed an improvement of 0.99 - 2.37\% in Macro F1 score using HTC-CLIP over the existing state-of-the-art models.
\end{abstract}

\end{frontmatter}

\section{Introduction}
\begin{figure}
\centering
    \includegraphics[width=0.45\textwidth]{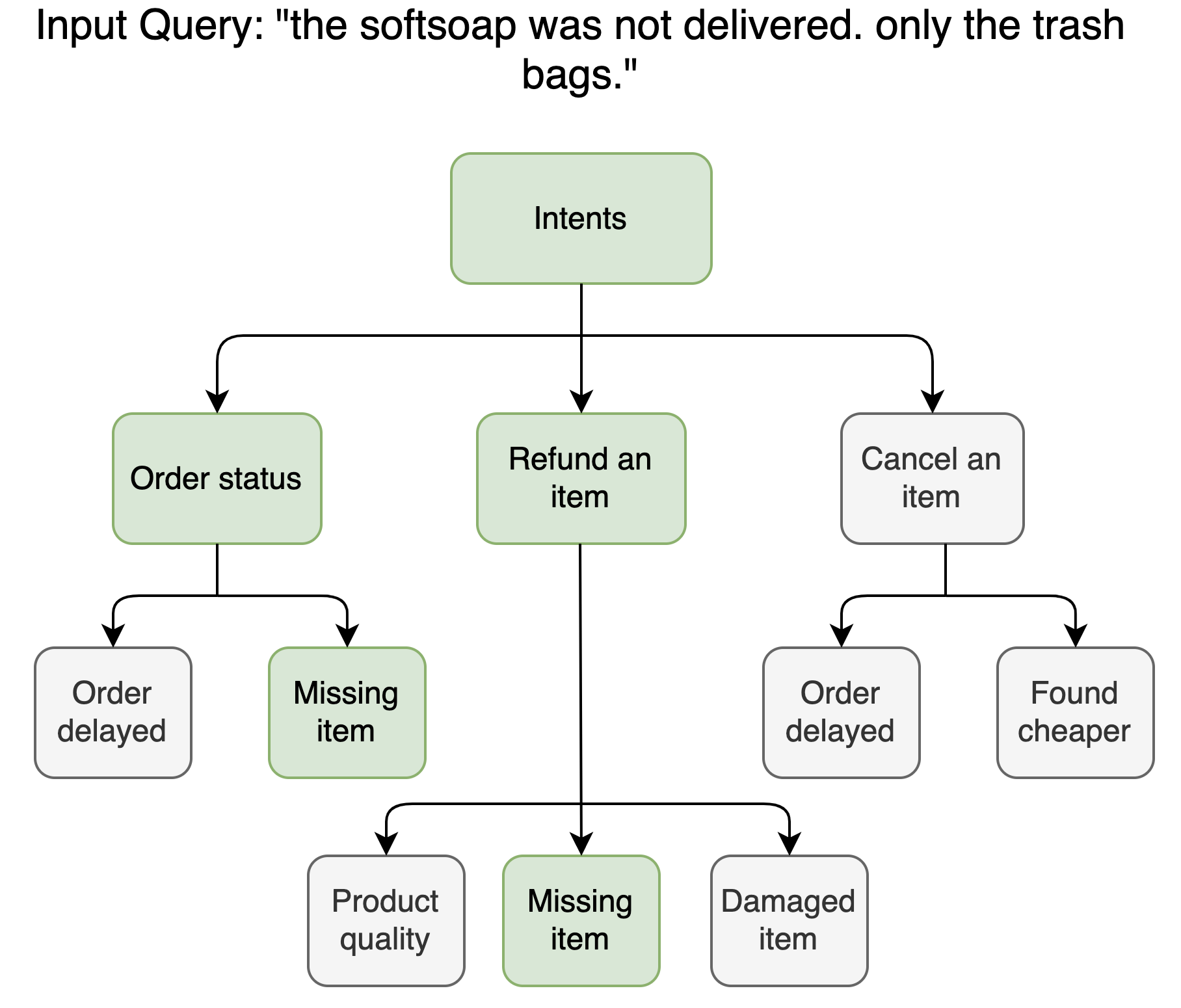}
    \caption{Input query is tagged with "missing items" from two different paths "order status" or "refund an item."}
    \label{dag}
\end{figure}

In the literature, the problem of categorizing text into a set of labels that are organized in a structured hierarchy is defined as Hierarchical Text Classification (HTC) \cite{silla2011survey, song2014dataless, meng2019weakly}.  HTC is a particular multi-label text classification (MLC) problem, where the classification result corresponds to one or more nodes of a taxonomic hierarchy. The class dependency in HTC is usually assumed to follow a hierarchical structure represented by a tree or a directed acyclic graph as shown in Figure \ref{dag}.

\begin{figure*}
\centering
    \includegraphics[width=1\textwidth]{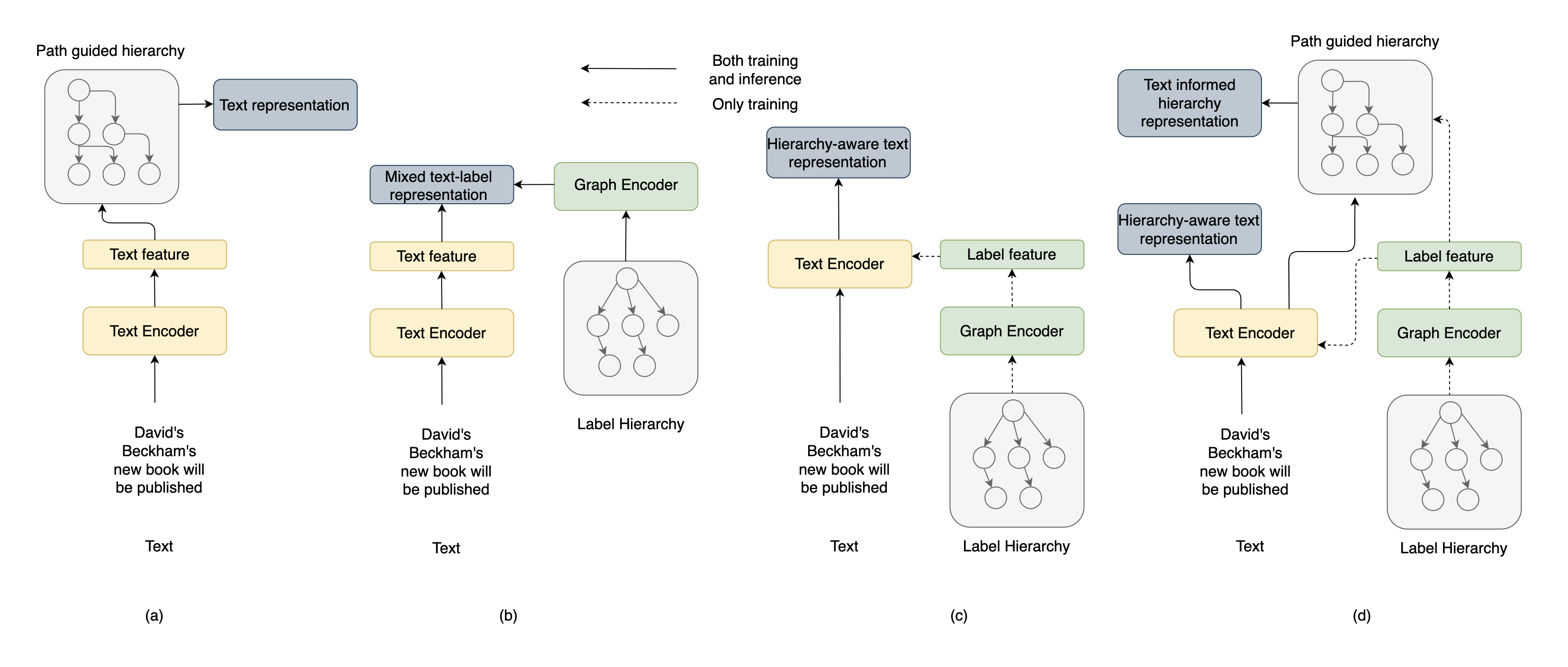}
    \caption{Different ways of introducing hierarchy information. (a) Previous work of modelling path-guided hierarchy, on top of text encoder \cite{wehrmann2018hierarchical}.  (b) Previous work of modelling text and labels separately and finding a mixed representation \cite{zhou-etal-2020-hierarchy, deng-etal-2021-htcinfomax}. (c) Previous work of incorporating hierarchy information into text encoder for a hierarchy-aware text representation \cite{wang-etal-2022-incorporating}. (d) Our work where the model learns two classifiers, one with text-encoded hierarchy and another with path-guided hierarchy, and pooled output from both classifiers is used during inference.}
    \label{htc_figure}
\end{figure*}

HTC approaches can be broadly classified as local approaches and global approaches. Algorithms that perform local learning attempt to discover the patterns that are present in regions of the class hierarchy, later combining the predictions to provide the ﬁnal classiﬁcation. The local approaches \cite{wehrmann2018hierarchical, shimura-etal-2018-hft, banerjee-etal-2019-hierarchical} exploit the parent and child hierarchy to overcome data imbalance in child node. Local approaches generally suffer from the error-propagation problem and are often computationally expensive. Global approaches for HTC, on the other hand, usually consist of a single classiﬁer capable of associating objects with their corresponding classes in the hierarchy at once \cite{10.1145/2487575.2487644, wu-etal-2019-learning}. Global approaches are usually less likely to capture local information from the hierarchy and eventually suffer from underfitting. \cite{10.1145/3019612.3019664} proposed a hybrid approach by combining local and global loss. Many researchers tried to encode text and label hierarchy separately and aggregate the two representations before being classified by a mixed feature \cite{zhou-etal-2020-hierarchy, deng-etal-2021-htcinfomax} (as shown in Figure \ref{htc_figure}b). Recently \cite{wang-etal-2022-incorporating} proposed a state-of-the-art model by encoding label hierarchy in a text encoder using contrastive learning during training time (as shown in Figure \ref{htc_figure}c). This model doesn't have a hierarchical encoder explicitly during inference time. The explicit label hierarchical encoder might learn information complement to the text encoder.

Therefore, in this paper, we introduce path-guided hierarchy using chained architecture and embedded hierarchy information in text using contrastive learning as shown in Figure \ref{htc_figure}d. We propose Hierarchical Text Classification using Contrastive Learning Informed Path guided hierarchy (HTC-CLIP), which exploits embedded hierarchy in the text  along with  path-guided hierarchy during inference time. We learn two sets of classifiers: one with linear layer and another with path-guided hierarchy, on top of hierarchy encoded text encoder. Graphormer \cite{NEURIPS2021_f1c15925} is used to embed hierarchy in BERT based text encoder \cite{wang-etal-2022-incorporating}. Linear and hierarchy classifiers learn different probability distributions. Therefore, during inference we use maximum pooling to get the final classifier probability.

The main contributions of our paper are summarised as follows:

\begin{itemize}
    \item We propose an architecture that can encode hierarchy information in text and explicitly learns path-guided hierarchy with contrastive learning.
    \item We introduce two sets of classifiers during training and pooling operation during inference to imitate ensemble behaviour.
    \item Experiments demonstrate that the proposed model achieves an improvement of 0.99-2.37\% in Macro F1 on two public datasets, WOS and NYT. 
\end{itemize}

\section{Related Work}
Work on HTC can broadly be categorized into local and global approaches \cite{wang-etal-2022-incorporating}. In local approaches, a classifier can be built per node, per parent, or per level. In global approaches only one classifier is built for the entire graph. \cite{banerjee-etal-2019-hierarchical} uses a local approach to build a classifier per label and transfers parameters of the parent model to the child model. \cite{10.1145/3019612.3019664, wehrmann2018hierarchical} use a hybrid approach to build a common classifier that optimizes global and local loss. \cite{10.1145/3178876.3186005} decomposes the hierarchy into subgraphs and conducts Text-GCN on n-gram tokens, whereas \cite{shimura-etal-2018-hft} applies CNN to utilize the data in the upper levels to contribute to categorization in the lower levels. 

In early global approaches, HTC is solved by reducing the problem to a flat multi-layer classification problem \cite{johnson-zhang-2015-effective}. Later approaches try to improve the MLC by embedding hierarchy information.  \cite{10.1145/2487575.2487644} introduced the label structure by recursive regularisation. Some works use deep learning based architecture to learn hierarchical structure of labels, by employing sequence to sequence network \cite{yang-etal-2018-sgm},  reinforcement learning \cite{mao-etal-2019-hierarchical}, meta-learning \cite{wu-etal-2019-learning} and capsule network \cite{peng2019hierarchical}. These models mainly focus on improving decoders based on the constraint of hierarchical paths. Later literature focused on encoding the hierarchy information using structure encoders. \cite{zhou-etal-2020-hierarchy} proposed GCN and LSTM-based hierarchical encoder and methods to fuse those embeddings with text encoder outputs. \cite{10.1145/3374217} extracts text features according to different hierarchy levels. \cite{deng-etal-2021-htcinfomax} introduces information maximization to constrain label representation learning. \cite{chen-etal-2021-hierarchy} views the problem as semantic matching and tries BERT as a text encoder. All these works tried to model structure encoder and text encoder separately and later create mixed representations for classification. Recently \cite{wang-etal-2022-incorporating} has shown that infusing hierarchy information in text encoder using Graphomer can further improve the state-of-the-art on three HTC datasets.

\section{Problem Definition}
Hierarchical Text Classification (HTC) is the task of classifying input text $x = \{x_1, x_2, ... , x_n\}$ to subset $y$ of label set Y, where n is the number of tokens in $x$. Size of label set Y is $\lvert C \rvert$. The label hierarchy mainly contains a tree-like structure and a directed acyclic graph (DAG) structure. We formulate label hierarchy as DAG, G = (Y, E), where node set Y is labels and edge set E denotes the relation between parent and child node. Since a non-root label
of HTC has one and only one parent, the label hierarchy can be converted to a tree-like hierarchy. Each sample $x$ corresponds to a subset  $y$ that includes multiple
classes. Those corresponding classes belong to either one or more sub-paths in the hierarchy as shown in Figure \ref{dag}.

\section{Methodology}
In this section, we will describe the proposed HTC-CLIP in detail. Figure \ref{hnlu} shows the overall architecture of the model.

\subsection{Text Encoder}
\label{le}
BERT \cite{devlin-etal-2019-bert} is one of the state-of-the-art encoders for text. We use it as the text encoder similar to  \cite{chen-etal-2021-hierarchy, wang-etal-2022-incorporating}. BERT uses WordPiece algorithm to tokenize the input text. Given an input text $x$, it gets tokenized into sequence of tokens $x_1, x_2,... x_{n-2}$. BERT adds [CLS] and [SEP], two special tokens indicating the beginning and the end of the sequence. Therefore, the final sequence of tokens of length $n$ is represented as:

\begin{equation}
    x = \{\textnormal{[CLS]}, x_1, x_2, ..., x_{n-2}, \textnormal{[SEP]}\} 
    \label{eq1}
\end{equation}

BERT encodes input tokens and outputs encodings corresponding to each token. We use encoding corresponding to [CLS] token as input to the path-guided hierarchy encoder and linear classifier.

\begin{equation}
    P = \textnormal{BERT}(x)
\end{equation}

Where, $P$ $\in \mathbb{R}^{n \times d_h}$, $d_h$ is hidden dimension and $n$ is number of tokens. Hidden state corresponding to [CLS] token is $p = P_{[CLS]}$.

\subsection{Hierarchy Encoder}
\cite{wang-etal-2022-incorporating} proposed a contrastive learning-guided hierarchy in text encoder and \cite{wehrmann2018hierarchical} suggested a path-guided hierarchy to improve the HTC task. Experiments suggest that probability class distributions learnt by these two methods are complementary to each other. In order to leverage the representation learnings of both HTC approaches, we have proposed a hybrid method capable of \textbf{simultaneously learning two sets of classifiers, one using path-guided hierarchy classifier and the other using linear classifier on top of hierarchy encoded text using contrastive learning}  as shown in Figure \ref{hnlu}.


\subsubsection{Path Guided Hierarchy Classifier}
\label{peh}
Path-encoded hierarchy helps in forcing the hierarchy structure between levels. \cite{wehrmann2018hierarchical, 10.1145/3019612.3019664} created the hierarchy on top of text encoder. We have proposed the configuration of linear layers with ReLU activation to learn better hierarchy relations between children and parents.

Formally, let $p \in \mathbb{R}^{d_h}$ be the pooled output (hidden state corresponding to [CLS] token) from BERT text encoder as described in section \ref{le}, $C_h$ be the set of classes of the $h^{th}$ hierarchical level, $\lvert H \rvert$ the total number of hierarchical levels, and  $\lvert C \rvert$ the total number of classes. Let $A_1^P$ denote the activations for the first level using BERT pooled output, given by:

\begin{equation}
    A_1^P =  \phi(W_1^P p + b_1^P)
\end{equation}

where $W_1^P \in \mathbb{R}^{C_1 \times d_h}$ is a weight matrix and $b_1^P \in \mathbb{R}^{C_1 \times 1}$ is the bias vector, which are the parameters for learning level $1$ classifier based on text and hierarchy encoded pooled output, and $\phi$ is a non-linear activation function, in our work we have used ReLU. Similarly, activation $A_h^P$ for level $h$ using pooled output is given by:
\begin{equation}
    A_h^P =  \phi(W_h^P p + b_h^P)
\end{equation}

where $W_h^P \in \mathbb{R}^{C_h \times d_h}$ is a weight matrix and $b_h^P \in \mathbb{R}^{C_h \times 1}$ is the bias vector.

We have introduced a linear layer, that maps activation from level $h-1$ to $k$ hidden neurons, and then a linear layer to map $k$ activation to $C_h$ classes at level $h$. These hidden layers help in learning a better representation of the parent and child nodes relationship. Let $A_h^{h-1}$ denote activation for the $h$ level using $h-1$ level activation.
\begin{align}
    A_k^{h-1} &=  \phi(W_k^{h-1} A_{h-1}^P + b_k^{h-1}) \label{eq:one} \\
    A_h^{h-1} &=  \phi(W_h^k A_k^{h-1} + b_h^k) \label{eq:two}
\end{align}

where $W_k^{h-1} \in \mathbb{R}^{k \times C_{h-1}}$, $W_h^k \in \mathbb{R}^{C_h \times k}$  are weight matrices and $b_k^{h-1} \in \mathbb{R}^{k \times 1}$, $b_h^k \in \mathbb{R}^{C_h \times 1}$ are the bias vectors.

Let $A_h$ be the final activation for level $h$, which is sum of activation based on pooled output, $A_h^p$ and activation based on previous level $A_h^{h-1}$. For level 1, $A_1^0$ is a vector of all ones.

\begin{equation}
    A_h = A_h^P \oplus A_h^{h-1}
\end{equation}

Following previous work \cite{wang-etal-2022-incorporating}, we flatten the hierarchy for multi-label classification therefore the output of each classifier is equal to the total number of classes $\lvert C \rvert$.  Hence, all activations from each level are concatenated and passed through sigmoidal activation to get class probabilities, $P_c$ of path-guided hierarchy classifier.

\begin{equation}
    P_c = \sigma(A_1 \odot A_2 \odot \cdots  \odot A_{\lvert H \rvert})
\end{equation}

\subsubsection{Positive Sample Generation for Contrastive Learning}
\label{psg}
The goal for the positive sample generation is to keep a fraction of tokens while retaining the labels. Given a token sequence as Equation \ref{eq1}, the token embedding of BERT is defined as:

\begin{equation}
     {e_1, e_2, ..., e_n} = \mathrm{BERT\_emb}(x)
\end{equation}
The scale-dot attention weight between token embedding and label feature is first calculated to
determine the importance of a token on a label, 

\begin{equation}
    q_i = e_iW_Q, k_j = l_jW_k, A_{ij} = \frac{q_ik^T_j}{\sqrt{d_h}}
\end{equation}

The query and key are token embeddings and label features respectively, and $W_Q \in R^{d_h \times d_h}$ and $W_K \in R^{d_h \times d_h}$ are two weight matrices. Thus, for a certain $x_i$, its probability of belonging to label $y_j$ can be normalized by a Softmax function.

Next, given a label $y_j$ , we can sample key tokens from that distribution and form a positive sample $\hat{x}$. To make the sampling differentiable, we replace the Softmax function with Gumbel-Softmax \cite{jang} to simulate the sampling operation:

\begin{equation}
    P_{ij} = gumbel\_softmax(A_{i1}, A_{i2}, ..., A_{ik})_j
\end{equation}

Notice that a token can impact more than one label, so we do not discretize the probability as one-hot vectors in this step. Instead, we keep tokens for positive examples if their probabilities of being sampled exceed a certain threshold $\gamma$, which can also control the fraction of tokens to be retrained. For multi-label classification, we simply add the
probabilities of all ground-truth labels and obtain the probability of a token $x_i$ regarding its ground-truth label set y as:

\begin{equation}
    P_i = \sum_{j \in y} P_{ij}
\end{equation}

Finally, the positive sample $\hat{x}$ is constructed as:

\begin{equation}
    \hat{x} = \{x_i ~ if ~ P_i > \gamma ~ else ~ \mathbf{0}\}
\end{equation}

where \textbf{0} is a special token that has an embedding of all zeros so that key tokens can keep their positions. The select operation is not differentiable, so we implement it differently to make sure the whole model can be trained end-to-end.
\begin{figure*}
\centering
    \includegraphics[width=1\textwidth]{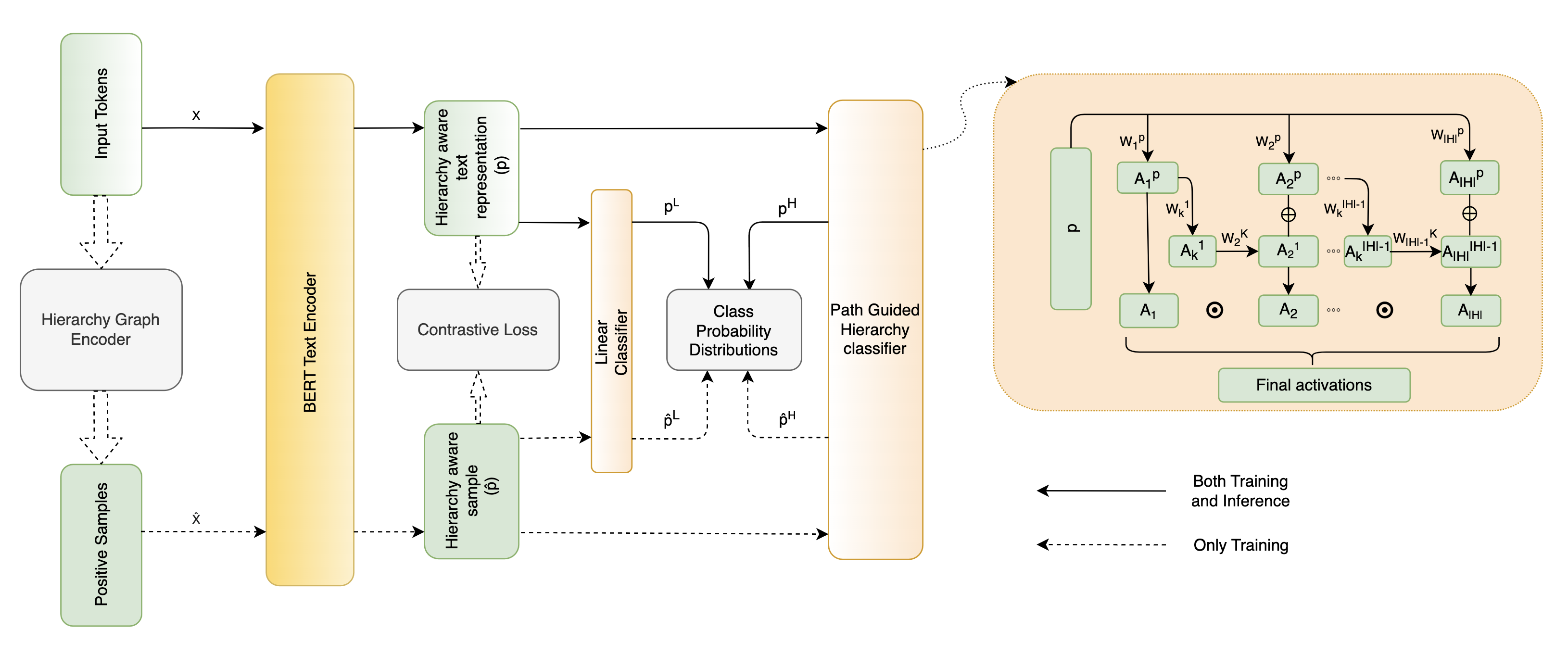}
    \caption{HTC-CLIP architecture. It contains a BERT based encoder, which generates a hierarchy-aware text representation and a positive sequence output representation. Both representations are passed through two classifiers (linear classifier and path-guided hierarchy classifier) to minimize binary cross-entropy loss, which helps in generating text-aware hierarchy representations.}
    \label{hnlu}
\end{figure*}
\subsubsection{Contrastive Learning for Text Encoder}
Contrastive loss helps in bringing token sequence and their positive counterpart representation closer, and the examples which are not from the same pair will be moved farther away. Encoding hierarchy information in text encoder using contrastive loss is proven to improve the result for HTC \cite{wang-etal-2022-incorporating}, we have used the same implementation for contrastive learning for text encoder. First, positive samples ($\hat{x}$) are generated for contrastive loss as discussed in \ref{psg}. The positive sample is fed to the same BERT as the original one in section \ref{le}.

\begin{equation}
    \hat{P} = \textnormal{BERT}(\hat{x})
\end{equation}

We get a sequence representation $\hat{p}$ with the first token corresponding to [CLS] before being classified.

With a batch of N hidden states of positive pairs ($p_i, \hat{p_i}$), with non-linear activation ReLU:
\begin{equation}
\begin{aligned}
c_i &= W_2 \textnormal{ReLU} (W_1 p_i)\\
\hat{c_i} &= W_2 \textnormal{ReLU} (W_1 \hat{p_i})\\
\end{aligned}
\end{equation}

where $W_1 \in \mathbb{R}^{d_h \times d_h}$ , $W_2 \in \mathbb{R}^{d_h \times d_h}$. 

For a batch of size N, we generate N positive examples so there will be total 2N pairs. For a given utterance there will be 2 positive pairs and remaining 2(N-1) pairs will be negative examples. Thus, in a batch of size N with 2N examples \textbf{Z} = $\{z \in \{c_i\} \cup \{\hat{c_i}\}\}$, we compute the NT-Xent loss \cite{pmlr-v119-chen20j} for $z_m$ as:

\begin{equation}
    L_m^{con} = -log \frac{exp(sim(z_m,\mu(z_m)/\tau)}{\sum_{i=1,i \neq m}^{2N} exp(sim(z_m, z_i)/\tau)} 
\end{equation}

where sim is the cosine similarity function as:
\begin{equation}
    sim(u, v) = {u \cdot v}/{\lVert u \rVert}{\lVert v \rVert}
\end{equation}

and $\mu$ is a matching function as:

\begin{equation}
    \mu(z_m) = 
    \begin{cases}
    c_i,& \text{if } z_m = \hat{c_i}\\
    \hat{c_i},              & \text{if} z_m =  c_i
\end{cases}
\end{equation}

$\tau$ is a temperature hyperparameter.

The total contrastive loss is the mean loss of all examples:
\begin{equation}
    L_{con} = \frac{1}{2N} \sum_{m=1}^{2N} L_m^{con}
\end{equation}

\begin{table*}

   \centering
    \begin{tabular}{c|c|c|c|c|c|c}
    \hline
         Dataset  & |L| &  Depth & Avg(|L\textsubscript{i}|) &Train  &  Val &   Test\\
         \hline
         
         WOS Dataset & 141 &  2  &  2.0  &  30,070 &   7,518 &   9,397\\
         
       
       NYT Dataset &  166  &  8  &  7.6 &   23,345  &  5,834 &   7,292\\
        \hline
         
    \end{tabular}
    \caption{Statistics of three datasets for hierarchical multi-label text classification. |L|: Number of target classes. Depth: Maximum level of hierarchy. Avg(|L\textsubscript{i}|): Average Number of classes per sample. Train/Val/Test: Size of train/validation/test set. }
    \label{tab:datasetdetails}
\end{table*}

\subsection{Contrastive Learning Informed Classifiers}
\label{clic}
To improve the representation of linear classifier and path-guided hierarchical classifier, the constructed positive sample BERT representation $\hat{p_i}$ is passed through each classifier separately. 

The probability of text representation $\hat{p_i}$ on label j is:
\begin{align}
    \hat{p}^{L}_{ij} &= \sigma(\text{Linear}(\hat{p_i}))_j  \label{a} \\
                 &=\sigma(W_L \cdot \hat{p_i} + b_L)_j \\ 
    \hat{p}^{H}_{ij} &= \sigma(\text{Hierarchical}(\hat{p_i}))_j \label{aa}
\end{align}

Where $W_L \in \mathbb{R}^{ \lvert C \rvert \times d_h}$ , $b_L \in \mathbb{R}^{ \lvert C \rvert \times 1}$. $\sigma$ is sigmoid function. The hierarchical classifier is explained in section \ref{peh}.

Positive sample representation, $\hat{p_i}$ learned using contrastive loss tunes classifiers' weights during training using binary cross entropy loss as explained in the next section. The  path-guided classifier's weights tuning helps in encoding text representation in a path-guided network.

\subsection{Classification  and Objective Function}
\label{bce}
Similar to positive constructed sequence output ($\hat{p_i}$), the hidden feature ($p_i$)  from the text encoder is fed into a linear and hierarchical classifier. Outputs from classifiers are fed into sigmoid ($\sigma$) activation to calculate class probability distribution.

The probability of text representation $p$ on label $j$ is:
\begin{align}
    p_{ij}^{L} &= \sigma(\text{Linear}(p_i))j\\
    p_{ij}^{H} &= \sigma(\text{Hierarchical}(p_i))j 
\end{align}

For multi-label classification, we use the binary cross-entropy loss function for text $i$ on label $j$,

\begin{align}
    L^C_{ij} = -y_{ij} \log(p_{ij}&) -  (1 - y_{ij})\log(1- p_{ij}) \label{b}\\
    L^C = & \sum_{i=1}^N \sum_{j=1}^k L_{ij}^C \label{c}
\end{align}

where $y_{ij}$ is the ground truth. $L^C$ will be $L^C_{L}$, $L^C_{H}$ for $p_{ij}$ as $p_{ij}^{L}$, $p_{ij}^{H}$ respectively.

The classification loss of the positive sample representation $\hat{L}^C_{L}$, $\hat{L}^C_{H}$ can be calculated similarly by $\hat{p}_{ij}^{L}$, $\hat{p}_{ij}^{H}$ using Equation \ref{a}, \ref{aa} and \ref{c}.

The final loss function is the combination of classification loss of original text, classification loss of the constructed positive samples, and the contrastive learning loss:

\begin{equation}
    \begin{aligned}
    L & = L^C_{L} + L^C_{H} +\hat{L}^C_{L} + \hat{L}^C_{H} + \lambda L_{con} \label{loss}
    \end{aligned}
\end{equation}

where $\lambda$ is a hyperparameter controlling the weight of contrastive loss.
During testing, we use the text encoder and path-guided encoder for classification and the model degenerates to a BERT encoder with two classification heads. The maximum pool output of both the linear classifier and path-guided classifier is used as a class probability. This output is used for assigning classes for a given input text.

\begin{table*}[]
\small
   \centering
    \begin{tabular}{c c c c c c c}
    
    \hline
          \multirow{2}{*}{Model} & \multicolumn{2}{c}{WOS Dataset}  & \multicolumn{2}{c}{NYT Dataset} \\
        
          & Micro-F1 & Macro-F1 & Micro-F1 & Macro-F1\\
         
         \noalign{\vskip 0.04in} 
         \hline
         \noalign{\vskip 0.04in} 
     
         BERT (Our implement) &85.80& 79.20    &78.22 & 65.85\\
         BERT+HiAGM \cite{wang-etal-2022-incorporating} & 86.04 & 80.19  &    78.64  &  66.76\\
         BERT+HTCInfoMax \cite{wang-etal-2022-incorporating} & 86.30 & 79.97 &     78.75  &  67.31\\
         BERT+HiMatch \cite{chen-etal-2021-himatch}  & 86.70 & 81.06    & - & -\\
          HGCLR \cite{wang-etal-2022-incorporating} (Our implement) &87.01& 80.30 &78.34&66.22 \\

         HTC-CLIP (Ours) &\textbf{87.86}& \textbf{81.64}    &\textbf{79.22}& \textbf{68.59}\\
        
         \hline

    \end{tabular}
    \caption{Experimental results comparing the proposed model (HTC-CLIP) with other state-of-the-art models on WOS and NYT datasets. For a fair comparison, we implemented the same baseline as the BERT encoder. We cannot reproduce the BERT and HGCLR results reported in \cite{wang-etal-2022-incorporating}, so we report the results of our implementation of BERT and HGCLR. }
    \label{tab:results}
\end{table*}

\section{Experiments}
In order to assess the effectiveness of the proposed architecture and compare it against existing models, we have run several experiments.

\subsection{ Experiment Setup}
\textbf{Datasets and Evaluation Metrics} We performed experiments on Web-of-Science (WOS) \cite{WOS_dataset} and NYT \cite{Sandhaus2008New} datasets  for comparison and analysis. WOS dataset includes abstracts of published papers from Web of Science. NYT is an archive of manually categorized newswire stories. For WOS and NYT datasets we have followed the train/val/test distribution of \cite{wang-etal-2022-incorporating} and \cite{zhou-etal-2020-hierarchy}. Statistics of these datasets are listed in Table \ref{tab:datasetdetails}. 

\textbf{Evaluation Metrics} 
We measure the experimental results by Micro-F1 and Macro-F1. Micro-F1 is calculated from the overall precision and recall of all the instances, while Macro-F1 is equal to the average F1-score of labels. 

\textbf{Implementation Details}
We use bert-base-uncased from Transformers \cite{wolf-etal-2020-transformers, wang-etal-2022-incorporating} as the base architecture for text encoder. As \cite{wang-etal-2022-incorporating} suggested, for Graphormer, attention head is set to 8 and feature size $d_{h}$ to 768. We use 1 GPU of Nvidia A100 for computing. We set the batch size to 56. Similar to \cite{wang-etal-2022-incorporating}, we use Adam optimizer with a learning rate of 3x10\textsuperscript{-5}. The threshold $\gamma$ is set to 0.02 on WOS, 0.005 on NYT dataset. The loss weight $\lambda$ is set to 0.05 on WOS, 0.3 on NYT dataset. The temperature of the contrastive module is fixed to 1 \cite{wang-etal-2022-incorporating}. The hidden layer size ($k$) for the path-guided hierarchy is set to 128 for all three datasets. We implemented our model in PyTorch and trained end-to-end, and stopped training if the Macro-F1 does not increase for 6 epochs. We evaluated the test subset with the model having the best Macro-F1 on the validation subset.

\textbf{Comparison Models}
We compare the performance of our HTC-CLIP model with a few recent works on HTC as strong baselines. HGCLR \cite{wang-etal-2022-incorporating}, HiAGM \cite{zhou-etal-2020-hierarchy}, HTCInfoMax \cite{deng-etal-2021-htcinfomax} and HiMatch \cite{chen-etal-2021-himatch}. HiAGM applies soft attention to text features and label features for the mixed feature. HTCInfoMax improves HiAGM by regularizing the label representation with a prior distribution. HiMatch matches text representation with label representation in a joint embedding space and uses joint representation for classification. HGCLR directly embeds the hierarchy into a text encoder.  HGCLR is the state-of-the-art before our work. Except for HiMatch and HGCLR, all of the above approaches adopt TextRCNN \cite{lai-et-al} as text encoder therefore we used \cite{wang-etal-2022-incorporating} implementation of these  with BERT for a fair comparison. Results are shown in Table \ref{tab:results}.

\textbf{Weight Parameters Size}
Table \ref{tab:abparam} shows the size of weight parameters. We observe that the number of parameters due to the path-guided hierarchy network in our proposed model doesn't increase the weight parameters significantly compared to HGCLR.


         
         

\subsection{ Experiment Results}
Table \ref{tab:results} reports the performance of our approach against other methods. On WOS dataset, our proposed HTC-CLIP model shows 2.06\% and 2.44\% improvement on Micro-F1 and Macro-F1 respectively compared to BERT. Our HTC-CLIP model also achieves competitive improvement over HGCLR and Hi-Match in terms of both Macro-F1 and Micro-F1. We could not reproduce the BERT and HGCLR results reported in \cite{wang-etal-2022-incorporating}, so we report the results of our implementation of BERT and HGCLR. We ran the experiment 5 times on WOS dataset and the standard deviation for Micro-F1 is 0.26 and Macro-F1 is 0.17. The mean of Micro-F1 and Macro-F1 for HGCLR are statistically different from that for HTC-CLIP as per t-test. HGCLR mean is more than three standard deviations away from that of HTC-CLIP for both Micro and Macro F1 scores. This shows that in comparison  to the state-of-the-art model, our model's results are statistically better.


On NYT, our approach shows 1.00\% and 2.74\% improvement on Micro-F1 and Macro-F1 respectively compared to BERT and performs significantly better than previous methods on both of the above measurements. The results show that HTC-CLIP achieves consistent improvement on the performance of Hierarchical Text Classification among WOS and NYT datasets.

\begin{table}
\small
    \centering
    \begin{tabular}{c|c|c}
    \hline
          Model & WOS  &   NYT   \\
         \hline
         
         BERT(Our implement) & 110.9 & 	 	110.9 \\
        HGCLR(Our implement) & 	120.5 & 		120.6\\
        HTC-CLIP & 	120.7 & 		 120.8 \\

    \end{tabular}
    \caption{ Number of weight parameters in the proposed model with other state-of-the-art models on WOS  and NYT datasets. The numbers displayed in the table are in millions.}
    \label{tab:abparam}
\end{table}

\subsection{ Analysis}

In this section, we investigate the independent effect of each component in our proposed
model.

\begin{table}
\small
    \centering
    \begin{tabular}{c|c|c}
    \hline
         Ablation Models & Micro-F1  &Macro-F1    \\
         \hline
          \multicolumn{1}{c|}{BERT} & 85.80 & 79.20\\
         \hline
         HTC-CLIP & \textbf{87.58} & \textbf{80.88}\\
         -r.m. $\hat{L}^C_{H}$ & 86.65 & 80.09\\
         -r.m. $\hat{L}^C_{L}$ & 86.95 & 80.13\\
         -r.m. $\hat{L}^C_{H}$\& $\hat{L}^C_{L}$ & 86.18 & 79.16\\
         -r.m. l.c. and $\hat{L}^C_{H}$ & 86.70 & 79.75\\
         
    \end{tabular}
    \caption{ Performance of HTC-CLIP on the validation set of WOS after removal of BCE loss components from Equation \ref{loss}. $\hat{L}^C_{H}$ and $\hat{L}^C_{L}$ are BCE loss of positive sample output representation learned through a contrastive loss for linear and path-guided hierarchy classifiers respectively, l.c. stands for a linear classifier, r.m. stands for remove}
    \label{tab:ab1}
\end{table}

\subsubsection{Effect of Contrastive Learning on Classifiers}
To demonstrate the advantage of contrastive information, we tested our model with and without BCE loss of constructed positive sequence output representation on the WOS Dataset. We first removed $\hat{L}^C_{H}$ from Equation \ref{loss}, results are shown in Table \ref{tab:ab1}. We observed that Macro-F1 and Micro-F1 scores drop significantly after removing $\hat{L}^C_{H}$. BCE loss of positive sequence output learned through contrastive loss  helps the path-guided hierarchy classifier in learning the text representation in its weights (section \ref{clic}). We also removed the $\hat{L}^C_{L}$ and saw a similar trend of decreased Macro-F1 and Micro-F1 scores. As evident from  Table \ref{tab:ab1}, BCE loss of positive sequence output representation learned through contrastive loss  helps in tuning weights and boosts performance in both linear classifier as well as path-guided hierarchy classifier.

\begin{table}
    \centering
    \small
    \begin{tabular}{c|c|c}
    \hline
         Ablation Models & Micro-F1  &Macro-F1    \\
         \hline
          BERT & 85.80 & 79.20\\
         \hline
         HTC-CLIP & \textbf{87.58} & \textbf{80.88}\\
         -r.m. h.c. & 86.69 & 80.38\\
         -r.m.  l.c. & 87.16 & 80.11 \\
         -r.m. hidden layers from h.c. & 86.38  & 79.88\\

    \end{tabular}
    \caption{ Performance of HTC-CLIP on the validation set of WOS after removal of classifiers. r.m. stands for remove,  h.c. stands for path-guided hierarchy classifier and l.c. stands for the linear classifier.} 
    \label{tab:ab2}
\end{table}


\subsubsection{Effect of Path Guided Hierarchy}
To study the influence of path-guided hierarchy, we tested our model on the WOS Dataset after removing the path-guided hierarchy classifier. Without the path-guided hierarchy classifier, we found a drop in Macro-F1 and Micro-F1 scores. Results are shown in Table \ref{tab:ab2}. We also experimented with the modified version of path-guided hierarchy after removing the hidden layers (Equation \eqref{eq:one} and \eqref{eq:two}, Figure \ref{hnlu}). As shown in Table \ref{tab:ab2}, these layers contribute positively to the model and help in learning a better representation of the hierarchy.


\begin{table}
    \centering
    \small
    \begin{tabular}{c|c|c}
    \hline
         Ablation Models & Micro-F1  &Macro-F1    \\
         \hline
          BERT & 85.80 & 79.20\\
         \hline
         Only h.c. model & 87.16 & 80.11\\
         Only l.c. model & 86.69& 80.39\\
         Both l.c. \& h.c. with avg pool  & 87.35 & 80.69  \\
         Both l.c. \& h.c. with max pool &  87.01 & 80.42 \\

         Both l.c. \& h.c. with max pool & \multirow{2}{*}{\textbf{87.58} } &  \multirow{2}{*}{ \textbf{80.88}}\\
         {during inference (HTC-CLIP)}&
    \end{tabular}
    \caption{ Performance of HTC-CLIP on the validation set of WOS after changing some components. h.c. stands for path-guided hierarchy classifier and l.c. stands for linear classifier}
    \label{tab:ab3}
\end{table}


\subsubsection{Effect of Pooling Classifiers' Outputs}

We trained the path-guided hierarchy model and linear classifier model with text-encoded hierarchy using contrastive loss. In our model, we propose to use both path-guided hierarchy and linear classifier in a single architecture, and a method to combine the output probabilities during inference.
As seen from Table \ref{tab:ab3}, our proposed architecture works better than individually trained models. This shows that the individual models capture complementary information and can be effectively combined into one architecture to achieve improved performance.  We find that adding or taking the maximum of the probabilities of classifier outputs during training yields better results than both the individual models, but is not able to outperform our proposed architecture where we do max pooling during inference.


\section{Conclusion}

In this paper, we present hierarchical
text classification using contrastive learning-informed path-guided hierarchy (HTC-CLIP). The method combines the strength of two existing approaches: contrastive learning guided hierarchy in text encoder and path guided hierarchy. Our paper shows that the two previous approaches capture complementary information and can be effectively combined into one architecture to achieve improved performance. Our approach empirically achieves consistent improvements over the state-of-the-art on two public benchmark datasets. All of the components we designed are proven to be effective.

\bibliography{ecai}
\end{document}